\documentclass{article}

 \usepackage[final]{bdl_2018}

\usepackage[utf8]{inputenc} 
\usepackage[T1]{fontenc}    
\usepackage{hyperref}       
\usepackage{url}            
\usepackage{booktabs}       
\usepackage{amsfonts}       
\usepackage{nicefrac}       
\usepackage{microtype}      

\usepackage{amsmath,amscd,amssymb}
\usepackage{array}
\usepackage{bbold}
\usepackage{bm}
\usepackage{mathtools}
\usepackage{hyperref}

\usepackage[table]{xcolor}
\usepackage{graphicx}
\usepackage{subcaption}
\usepackage{tabu}

\newcommand{\bx}{{\bm{x}}}

\newcommand{\bz}{{\bm{z}}}

\newcommand{\NN}{\mathcal{N}}
\newcommand{\htw}{{\tilde{h}}}

\newcommand{\ba}{\mathbf{a}}
\newcommand{\bat}{\mathbf{a}^{\rm T}}

\newcommand{\by}{\mathbf{y}}
\newcommand{\byt}{\hat{\mathbf{y}}}

\newcommand{\bX}{\mathbf{X}}

\newcommand{\bUps}{\bm{\Upsilon}}

\newcommand{\bSig}{\bm{\Sigma}}
\newcommand{\bA}{\mathbf{A}}

\newcommand{\batz}{{\ba_0^{\rm T}}}
\newcommand{\baz}{{\ba_0}}
\newcommand{\bBz}{{\bB_0}}
\newcommand{\bBzt}{{\bB_0^{\rm T}}}
\newcommand{\bPhi}{\mathbf{\Phi}}

\newcommand{\bxt}{{\mathbf{x}^{\rm T}}}

\newcommand{\bXi}{\bm{\Xi}}

\newcommand{\bB}{\mathbf{B}}

\newcommand{\bY}{\mathbf{Y}}

\title{Closed Form Variational Objectives For Bayesian Neural Networks with a Single Hidden Layer}

\author{
 Martin Jankowiak  \hspace{1.5mm}  \texttt{jankowiak@uber.com} \\
  Uber AI Labs \hspace{1.5mm} San Francisco, CA }

\begin{document}

\maketitle

\begin{abstract}
In this note we consider setups in which variational objectives for Bayesian neural networks can be computed in closed form.
In particular we focus on single-layer networks in which the activation function is piecewise polynomial (e.g.~ReLU).
In this case we show that for a Normal likelihood and structured Normal variational distributions one can compute a variational
lower bound in closed form. In addition we compute the predictive mean and variance in closed form. Finally, we also show how to compute approximate lower bounds for other likelihoods (e.g.~softmax classification). In experiments
we show how the resulting variational objectives can help improve training and provide fast test time predictions.
\end{abstract}


\section{Introduction}

In recent years significant effort has gone into developing flexible probabilistic models for the supervised setting.
These include, among others, deep gaussian processes \citep{damianou2013deep} as 
well as various approaches to Bayesian neural networks \citep{peterson1987mean,mackay1992practical,hinton1993keeping,graves2011practical,blundell2015weight,hernandez2015probabilistic}.
While neural networks promise considerable flexibility, scalable learning algorithms for Bayesian neural networks that can deliver robust uncertainty estimates remain elusive. While some of the difficulty stems from the inadequate (weight-space) priors that are typically used, much of the challenge can be traced to the difficulty of the inference problem itself. In the variational inference setting, this manifests itself in at least two ways. First, the need to restrict the variational family to a tractable class limits the fidelity of the approximate learned posterior. Second, nested non-linearities necessitate sampling methods during training, which can make for a challenging stochastic optimization problem, especially for 
wide, deep networks. In this work our goal is to make the stochastic optimization problem (somewhat) easier by integrating out some of the weights analytically. In the next section we focus on the regression case, leaving a discussion of other cases to the appendix.\footnote{Please refer to the appendix for a more detailed discussion of related work.}

\section{Regression Setup}

Consider a dataset $\{\bx_i, \by_i\}$ of size $N$ with inputs $\bx_i$ and outputs $\by_i$.
To simplify the notation, we consider the case where $\bx$ is $D$-dimensional and $\by$ is 1-dimensional.
We consider a neural network with a single hidden layer defined by the following computational flow:\footnote{We handle
bias terms by augmenting inputs to each neural network layer with an element equal to 1.}
\begin{equation}
\bx \rightarrow \bA \bx  \rightarrow g(\bA \bx)  \rightarrow \bB g(\bA \bx) 
\end{equation}
Here $g(\cdot)$ is the non-linearity, $\bA$ is of size $H \times D$ and $\bB$ is of size $1 \times H$, where $H$ is the number of hidden units.
We choose a Normal likelihood with precision $\beta$ and standard  Normal priors for the weights. 
Thus the marginal likelihood of the observed data is:
\begin{equation}
p(\bY | \bX) = \int d\bA d\bB \;p(\bA) p(\bB) \prod_i  p(\by_i | \bB g(\bA \bx_i), \beta) 
\end{equation}

\section{Variational Bound}

We consider a variational distribution of the form
\begin{equation}
\label{eqn:varform}
q(\bA, \bB) = q(\bB) \prod_{h=1}^H q(\ba_h)  \;\;\; {\rm with} \;\; \;
 q(\ba_h)  = \mathcal{N}(\ba_h | \baz_h, \bSig_{\ba_h}) \;\;\; {\rm and} \;\;\; q(\bB)  = \mathcal{N}(\bB | \bBz, \bSig_{\bB})
\end{equation}
where each component distribution is Normal. 
Since we treat each row of $\bA$ independently, 
the activations $\{ g(\bA \bx)_h \}$ are conditionally independent given an input $\bx$. With these assumptions
we can write down the following variational bound:
\begin{equation}
\log p(\bY | \bX) \ge \mathbb{E}_{q(\bA)q(\bB)} \left[ \sum_i \log p(\by_i | \bx_i, \bA, \bB) \right] - {\rm KL}(q(\bA) | p(\bA))  - {\rm KL}(q(\bB) | p(\bB)) 
\label{eqn:elbo}
\end{equation}
The KL divergences are readily computed. 
We now show that we can compute closed form expressions for the first term in Eqn.~\ref{eqn:elbo} (i.e.~the expected log likelihood) for certain non-linearities $g(\cdot)$. For
concreteness we consider the ReLU activation function, i.e.~$g(x) = {\rm max}(0, x) = \tfrac{1}{2}(x + |x|)$.
The expected log likelihood (ELL) for a single datapoint is given by
\begin{equation}
 {\rm ELL} = -\frac{\beta}{2} \mathbb{E}_{q(\bA)q(\bB)} \left[ (\by - \bB g(\bA \bx))^2\right] + \frac{1}{2}\left(\log \beta - \log 2 \pi  \right)
 \label{eqn:singleell}
\end{equation}
The expectation in Eqn.~\ref{eqn:singleell} becomes
\begin{equation}
\mathbb{E}_{q(\bA)q(\bB)} \Big[ \by^2 -\by\bB_{1h}\left(\bat_h \bx + |\bat_h \bx|\right) 
+ \tfrac{1}{4}\Sigma_{h, \htw} \bB_{1h}\bB_{1\htw}\left(\bat_h \bx + |\bat_h \bx|\right)\left(\bat_\htw \bx + |\bat_\htw \bx|\right)\Big]
\end{equation}
Massaging terms, the expected log likelihood for the full dataset is given by
\begin{equation}
{\rm ELL} = -\frac{\beta}{2}\sum_i \left[ (\by_i  -  \byt_i(\bx_i))^2 +{\rm var}_{\bA \bB}(\bx_i) \right] + \frac{N}{2}\left( \log \beta - \log 2 \pi\right)
\end{equation}
Here we have introduced the mean function $\byt(\bx) = \bB_0 \cdot \bPhi$ 
as well as the corresponding variance:
\begin{equation}
{\rm var}_{\bA \bB}(\bx) =  {\rm diag}(\bXi_\bB) \cdot (\bUps - \bPhi \odot \bPhi) + \bPhi^{\rm T}\bSig_\bB\bPhi 
\end{equation}
Note that $\byt(\bx)$ and ${\rm var}_{\bA \bB}(\bx)$ can be used at test time to yield fast predictive means and variances.
We have also defined the matrix $\bXi_\bB = \bSig_\bB + \bBz \bBzt$ and the $H$-dimensional vectors  
$\bPhi$ and $\bUps$:
\begin{equation}
\label{eqn:phiupsdef}
\Phi_h= \mathbb{E}_{q(\bA)} \left[ g(\bA \bx)_h \right] \qquad \qquad
\Upsilon_h= \mathbb{E}_{q(\bA)} \left[ g(\bA \bx)_h^2 \right] 
\end{equation}
The key quantities are the expectations in Eqn.~\ref{eqn:phiupsdef}. As we show in the appendix, these can be computed in closed form
for piecewise polynomial activation functions. The resulting expressions involve nothing more exotic than the error function.

\section{Experiments}

We present a few experiments that demonstrate how our approach can be folded into larger probabilistic
models. Note that our focus here is on how (partial) analytic control can help training (Sec.~\ref{sec:variance}-\ref{sec:vae}) and
prediction (Sec.~\ref{sec:resnet}) and \emph{not} the suitability of Bayesian neural networks for particular tasks or datasets.
Please refer to the appendix for details on experimental setups.

\subsection{Variance Reduction}
\label{sec:variance}

We train a Bayesian neural network with two hidden layers on a regression task and compute the gradient variance during training. As can be seen from Table~\ref{table:varred}, Rao-Blackwellizing the two weight matrices closest to the outputs reduces the variance, especially for the covariance parameters. As the weight matrices we integrate out get larger, this variance reduction becomes more pronounced.
\newcommand{\layonemeanearlyanal}{\small 8.6}
\newcommand{\layonemeanlateanal}{\small1.75} 
\newcommand{\layonescaleearlyanal}{\small$6\!\times\! 10^{-4}$}
\newcommand{\layonescalelateanal}{\small$1\!\times\! 10^{-4}$}
\newcommand{\laytwomeanearlyanal}{\small3.2}
\newcommand{\laytwomeanlateanal}{\small0.2}
\newcommand{\laytwoscaleearlyanal}{\small$2\!\times\! 10^{-6}$}
\newcommand{\laytwoscalelateanal}{\small$2\!\times\! 10^{-6}$}
\newcommand{\laythreemeanearlyanal}{\small227}
\newcommand{\laythreemeanlateanal}{\small259}
\newcommand{\laythreescaleearlyanal}{\small$1\!\times\! 10^{-5}$}
\newcommand{\laythreescalelateanal}{\small$3\!\times\! 10^{-8}$}
\newcommand{\layonemeanearlysamp}{\small19.7}
\newcommand{\layonemeanlatesamp}{\small2.6}
\newcommand{\layonescaleearlysamp}{\small$7\!\times\! 10^{-4}$}
\newcommand{\layonescalelatesamp}{\small$1\!\times\! 10^{-4}$}
\newcommand{\laytwomeanearlysamp}{\small10.1}
\newcommand{\laytwomeanlatesamp}{\small0.4}
\newcommand{\laytwoscaleearlysamp}{\small$2\!\times\! 10^{-4}$}
\newcommand{\laytwoscalelatesamp}{\small$1\!\times\! 10^{-4}$}
\newcommand{\laythreemeanearlysamp}{\small704}
\newcommand{\laythreemeanlatesamp}{\small518}
\newcommand{\laythreescaleearlysamp}{\small0.03}
\newcommand{\laythreescalelatesamp}{\small$3\!\times\! 10^{-3}$}
\begin{table}[h!]
\begin{center}
    \begin{tabu}{|c|[1pt]c|c|c|c|c|c|}    \hline
     \cellcolor[gray]{0.66} & \multicolumn{3}{c|}{\small Upon initialization \cellcolor[gray]{0.77}} 
     & \multicolumn{3}{c|}{\small Late in training \cellcolor[gray]{0.77}}  \\  \hline
   \cellcolor[gray]{0.66} & \small First Layer \cellcolor[gray]{0.95}& \cellcolor[gray]{0.95}\small Second Layer  &\cellcolor[gray]{0.95}\small Final Layer  
   \cellcolor[gray]{0.95} & \small First Layer \cellcolor[gray]{0.95}& \cellcolor[gray]{0.95}\small Second Layer  &\cellcolor[gray]{0.95}\small Final Layer   \\  \tabucline[1pt]{-}
   \small Analytic \cellcolor[gray]{0.95}& \layonemeanearlyanal \;/ \layonescaleearlyanal 
   & \laytwomeanearlyanal \;/ \laytwoscaleearlyanal & \laythreemeanearlyanal \;/ \laythreescaleearlyanal   
   & \layonemeanlateanal \;/ \layonescalelateanal & \laytwomeanlateanal \;/ \laytwoscalelateanal 
   & \laythreemeanlateanal \;/ \laythreescalelateanal \\ \hline
   \small Sampling \cellcolor[gray]{0.95}& \layonemeanearlysamp \;/ \layonescaleearlysamp
   & \laytwomeanearlysamp \;/ \laytwoscaleearlysamp & \laythreemeanearlysamp \;/ \laythreescaleearlysamp   
   & \layonemeanlatesamp \;/ \layonescalelatesamp & \laytwomeanlatesamp \;/ \laytwoscalelatesamp 
   & \laythreemeanlatesamp \;/ \laythreescalelatesamp \\ \hline
    \end{tabu}
\end{center}
     \caption{Mean gradient variances for the network in Sec.~\ref{sec:variance}. The first number in each cell corresponds to gradients w.r.t.~weight means and the second to gradients w.r.t.~(log root) variances.}
     \label{table:varred}
\end{table}

\subsection{VAE with a Bayesian Decoder}
\label{sec:vae}

We train a VAE \citep{kingma2013auto,rezende2014stochastic} with a Normal likelihood on a continuous-valued dataset. For the decoder we use a Bayesian neural network with a single hidden layer.\footnote{Alternatively, we can think of this
as the neural network analog of the deep latent variable model in ref.~\citep{dai2015variational}.}
We train three model and inference variants and report test log likelihoods in Table~\ref{table:vae}.
Apart from the first variant (V1), all variants make use of 
a Bayesian neural network as the decoder. Variants V2 and V3 differ in which weights are sampled during training.\footnote{More precisely, we always use the `local reparameterization trick' \citep{kingma2015variational}
and never sample weights directly.} In V2 the weights before
the non-linearity are sampled, while in V3 no weights are sampled. While the test log likelihoods in Table~\ref{table:vae} do not differ dramatically, we see evidence that: i) a Bayesian decoder can be useful in this setting; and ii) integrating out weights can help us train a better model.

\begin{table}[h!]
\begin{center}
    \begin{tabu}{|c|[1pt]c|c|c|}    \hline
   \cellcolor[gray]{0.77} & V1 \cellcolor[gray]{0.95}& \cellcolor[gray]{0.95}V2  &\cellcolor[gray]{0.95}V3 (this work) \\  \tabucline[1pt]{-}
   Bayesian Decoder \cellcolor[gray]{0.95}& No &  Yes & Yes \\ \hline
   Sampling \cellcolor[gray]{0.95}& $\bz$ only & $\bz$ and some weights & $\bz$ only \\ \hline
   Test LL \cellcolor[gray]{0.95}& -107.16 & -107.20 & \bf{-107.10} \\ \hline
    \end{tabu}
\end{center}
     \caption{Test log likelihoods for the VAE experiment in Sec.~\ref{sec:vae}. Higher is better.}
     \label{table:vae}
\end{table}

\subsection{Fast Prediction}
\label{sec:resnet}

We train a Bayesian neural network on ImageNet \citep{russakovsky2015imagenet}.\footnote{While our analytic results can be used to form approximate variational objectives (or, alternatively, control variates) in the classification setting (see Appendix) here we sample the weights during training.} Specifically we place a prior on the two weight matrices closest to the softmax output. We then compute classification accuracies on the test set using two methods: i) Monte Carlo; and ii) a deterministic approximation using the analytic results described above (see Appendix for details). As can be
seen from Fig.~\ref{fig:resnet}, a large number of samples must be drawn before the MC estimator reaches the performance of the deterministic approximation. Indeed even with $512$ samples the deterministic approximation outperforms MC on top-5 accuracy.

\begin{figure}
\centering
\begin{subfigure}{.5\textwidth}
  \centering
  \includegraphics[width=.98\textwidth]{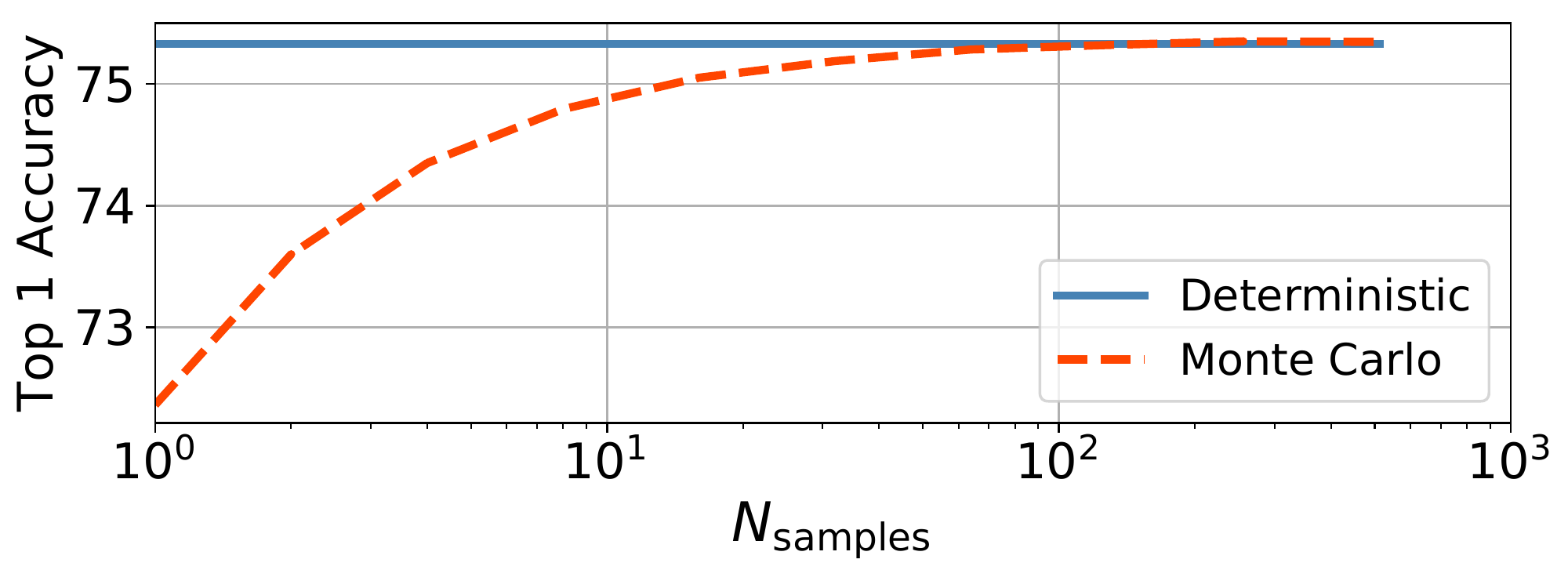}
  \label{fig:sub1}
\end{subfigure}%
\begin{subfigure}{.5\textwidth}
  \centering
  \includegraphics[width=.98\textwidth]{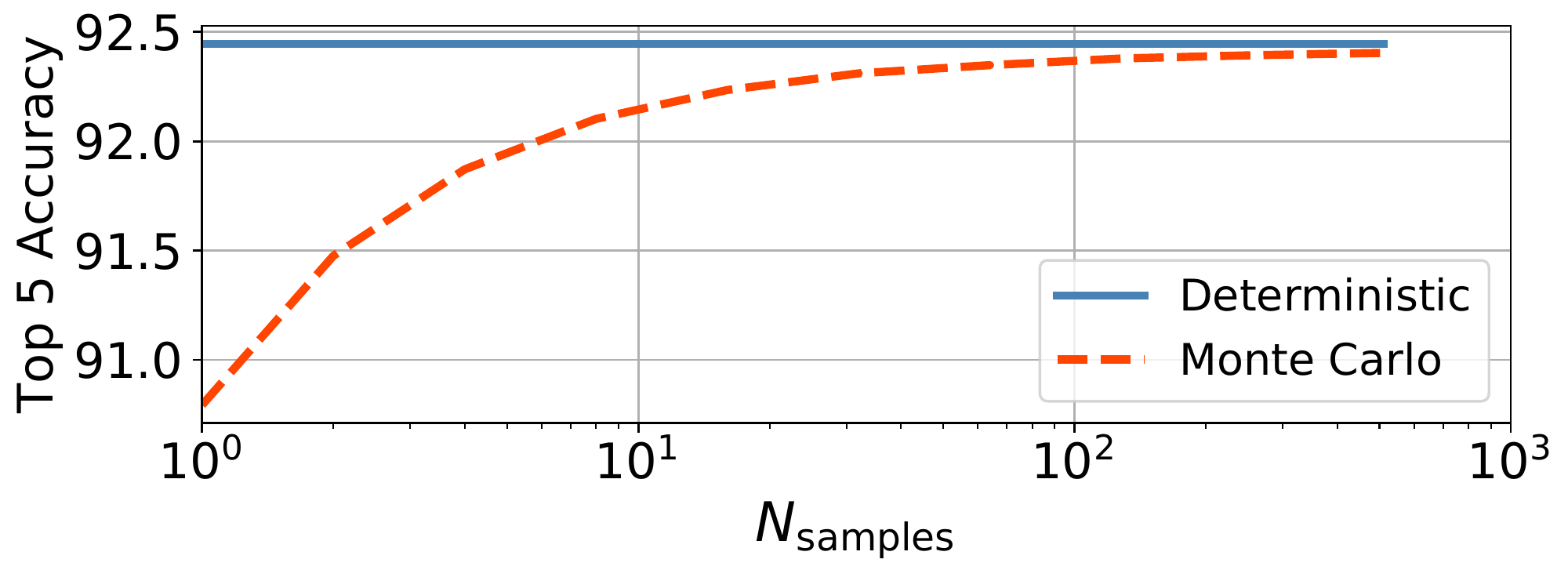}
  \label{fig:sub2}
\end{subfigure}
\caption{We compare the performance of a Monte Carlo estimate of classification accuracy to a deterministic approximation. 
{\bf Left:} Top-1 accuracy. {\bf Right:} Top-5 accuracy.  See Sec.~\ref{sec:resnet} for details.}
\label{fig:resnet}
\end{figure}

\section{Discussion}

The approach developed here is expected to be most useful when integrated into larger Bayesian neural networks setups. 
It would be of particular interest to combine this approach with the class of priors described in \citep{karaletsos2018probabilistic}, the deterministic approximations in \citep{wu2018fixing}, or Normal variational distributions with flexible conditional dependence like those in \citep{louizos2017multiplicative}. Finally, our analytic results could be useful in the context of other classes of non-linear probabilistic models.

\bibliographystyle{abbrv} 
\bibliography{closed_form}

\section{Appendix}

The main goal of this appendix is to show how to compute the necessary expectations in
Eqn.~\ref{eqn:phiupsdef} for piecewise polynomial non-linearities.
Instead of presenting a (unwieldy) master formula for the general case, we proceed step by step and show how the computation is done in a few cases of increasing complexity. We begin with a basic ReLU integral.

\subsection{ReLU Mean Function}

We first consider the mean function for the ReLU activation function $g(x) = {\rm max}(0, x) = \tfrac{1}{2}(x + |x|)$, i.e.~we would like to compute the following expectation:
\begin{equation}
\label{eqn:relumean}
\mathbb{E}_{q(\ba)} \left[\bat \bx + |\bat \bx|  \right] =
\mathbb{E}_{q(\ba)} \left[\bat \bx  \right] +\mathbb{E}_{q(\ba)} \left[\ |\bat \bx|  \right] 
\end{equation}
where $q(\ba)= \mathcal{N}(\ba | \ba_0, \bSig_\ba)$. The first expectation in Eqn.~\ref{eqn:relumean} is elementary.
For the second expectation note that,
since $\bat \bx \sim \NN(\batz \bx,\bxt \bSig_\ba \bx)$, the expectation can be transformed to a one-dimensional integral
\begin{equation}
\label{eqn:relumean2}
\mathbb{E}_{q(\ba)} \left[ |\bat \bx|  \right] = \mathbb{E}_{y \sim \NN(\batz \bx,\bxt \bSig_\ba \bx)} \left[ |y|  \right] 
\end{equation}
that can readily be computed in terms of the error function. We do not do so here, 
however, because in subsequent derivations we will find an 
alternative strategy---namely to make use of a particular integral representation for the absolute value function---to be more
convenient. Thus before we give an explicit formula for Eqn.~\ref{eqn:relumean} we collect a few useful identities.

\subsection{Useful Integrals}

First we define the following (scalar) quantities, which we will make extensive use of throughout the appendix:
\begin{equation}
\begin{split}
\qquad \xi^2 \equiv \bxt \bSig_\ba \bx \qquad \qquad
\gamma \equiv \batz \bx
\end{split}
\end{equation}
We then have
\begin{equation}
\begin{split}
\mathbb{E}_{q(\ba)} \left[ \exp(i \bat \bx t )\right] = \exp(-\tfrac{1}{2} \xi^2 t^2 + i \gamma t)  
\end{split}
\end{equation}
and
\begin{equation}
\mathbb{E}_{q(\ba)} \left[ \bat \bx \exp(i \bat \bx t )\right] = \left( i \xi^2 t + \gamma \right)\exp(-\tfrac{1}{2} \xi^2 t^2 + i \gamma t) 
\end{equation}
as well as
\begin{equation}
\mathbb{E}_{q(\ba)} \left[ (\bat \bx)^2 \right] = \xi^2 + \gamma^2
\end{equation}

\noindent The  integral identity we make use of is:\footnote{Note that this identity is also used in \citep{li2009gaussian} to compute a related class of Gaussian integrals.}
\begin{equation}
\label{eqn:absrep}
|z| = \frac{2}{\pi} \int_0^{\infty} \frac{dt}{t^2} \left( 1 - \cos(zt)\right) = \frac{2}{\pi} \int_0^{\infty} \frac{dt}{t^2} \left( 1 - \frac{\exp(izt)+\exp(-izt)}{2}\right) 
\end{equation}
For reference we note that this identity can easily be derived by integrating by parts and
making use of the well-known sine integral:\footnote{The absolute value in Eqn.~\ref{eqn:absrep} is an immediate consequence of $\cos(zt) = \cos(-zt)$.}
\begin{equation}
\int_0^{\infty} \frac{\sin(t)}{t}dt=\tfrac{1}{2}\pi
\end{equation}

\subsection{ReLU Part II}
Combining the above identities we get:
\begin{equation}
\begin{split}
\label{eqn:omega_integral}
\mathbb{E}_{q(\ba)} \left[ |\bat \bx|  \right] &= \frac{2}{\pi} \int_0^{\infty} \frac{dt}{t^2} \Big( 1 - e^{-\tfrac{1}{2} \xi^2 t^2} \cos (\gamma t)\Big) \\
&= \sqrt{\frac{2}{\pi}} \xi e^{-\tfrac{1}{2} \frac{\gamma^2}{\xi^2}} + \gamma {\rm erf}\left(\frac{\gamma}{\sqrt{2} \xi}\right) \\
&\equiv \Omega(\xi, \gamma)
\end{split}
\end{equation}
and
\begin{equation}
\begin{split}
\mathbb{E}_{q(\ba)} \left[ (\bat \bx) |\bat \bx|  \right] &= \frac{2}{\pi} \int_0^{\infty} \frac{dt}{t^2} 
\Big( \gamma - \gamma e^{-\tfrac{1}{2} \xi^2 t^2} \cos (\gamma t) + \xi^2 t e^{-\tfrac{1}{2} \xi^2 t^2} \sin (\gamma t)\Big)\\
&= \gamma \left\{  \frac{2}{\pi} \int_0^{\infty} \frac{dt}{t^2} \Big( 1 - e^{-\tfrac{1}{2} \xi^2 t^2} \cos (\gamma t)\Big) \right\}  
+\xi^2 \frac{\partial}{\partial \gamma} \left\{\frac{2}{\pi} \int_0^{\infty} \frac{dt}{t^2} \Big( 1 - e^{-\tfrac{1}{2} \xi^2 t^2} \cos (\gamma t)\Big) \right\}\\
&= \gamma \left\{ \sqrt{\frac{2}{\pi}} \xi e^{-\tfrac{1}{2} \frac{\gamma^2}{\xi^2}} + \gamma {\rm erf}\left(\frac{\gamma}{\sqrt{2} \xi}\right)\right\} +
\xi^2 \frac{\partial}{\partial \gamma} \left\{ \sqrt{\frac{2}{\pi}} \xi e^{-\tfrac{1}{2} \frac{\gamma^2}{\xi^2}} + \gamma {\rm erf}\left(\frac{\gamma}{\sqrt{2} \xi}\right)\right\}\\
&= \sqrt{\frac{2}{\pi}} \gamma \xi e^{-\tfrac{1}{2} \frac{\gamma^2}{\xi^2}} + (\xi^2 + \gamma^2) {\rm erf}\left(\frac{\gamma}{\sqrt{2} \xi}\right) \\
&\equiv \Psi(\xi, \gamma)
\end{split}
\end{equation}
Thus we have all the ingredients to compute the expectations in Eqn.~\ref{eqn:phiupsdef}:
\begin{equation}
\begin{split}
\label{eqn:phiupsrelu}
&\Phi_h= \mathbb{E}_{q(\bA)} \left[ g(\bA \bx)_h \right] = \tfrac{1}{2}(\gamma_h + \Omega_h) \\
&\Upsilon_h= \mathbb{E}_{q(\bA)} \left[ g(\bA \bx)_h^2 \right] =\tfrac{1}{2}(\xi_h^2 + \gamma_h^2 + \psi_h)
\end{split}
\end{equation}
As stated in the main text, these expectations involve nothing more exotic than the error function.
Note that as $\gamma_h / \xi_h \to \infty$ only a small portion of probability mass is propagated through
the constant portion of the ReLU activation function. As such, in this limit we expect $\Phi_h \to \gamma_h$ and
$\Upsilon_h \to \gamma_h^2 + \xi_h^2$. It is easy to verify that this is indeed the case. Similarly, as
$\gamma_h/ \xi_h \to -\infty$, we have $\Phi_h, \Upsilon_h \to 0$.

\subsection{Other Non-linearities}

\subsubsection{Leaky ReLU}

We consider the `leaky' ReLU, which we define to be given by
\begin{equation}
g_\epsilon(x) =  {\rm max}(\epsilon x, x) = \tfrac{1}{2}((1+\epsilon)x + (1-\epsilon)|x|)
\end{equation}
for some $\epsilon>0$. In this case one finds:
\begin{equation}
\begin{split}
\label{eqn:phiups}
&\Phi_h= \mathbb{E}_{q(\bA)} \left[ g_\epsilon(\bA \bx)_h \right] = \tfrac{1}{2}((1+\epsilon)\gamma_h + (1-\epsilon)\Omega_h) \\
&\Upsilon_h= \mathbb{E}_{q(\bA)} \left[ g_\epsilon(\bA \bx)_h^2 \right] =\tfrac{1}{2}((1+\epsilon^2)(\xi_h^2 + \gamma_h^2) + (1-\epsilon^2)\psi_h)
\end{split}
\end{equation}

\subsubsection{Hard Sigmoid}

We consider the `hard sigmoid' non-linearity, which we define to be given by 
\begin{equation}
g_\alpha(x) = \tfrac{1}{2}(|x+\alpha| - |x-\alpha|)
\end{equation}
for a given constant $\alpha$.
The identity in Eqn.~\ref{eqn:absrep} can immediately be generalized to
\begin{equation}
\begin{split}
|z+z_0| &= \frac{2}{\pi} \int_0^{\infty} \frac{dt}{t^2} \Big( 1 - \cos(zt)\cos(z_0t)+\sin(zt)\sin(z_0t)\Big) \\
&= \frac{2}{\pi} \int_0^{\infty} \frac{dt}{t^2} \Big( 1 - \frac{1}{2} [ \exp(izt + i z_0t)+\exp(-izt-iz_0t))] \Big)
\end{split}
\end{equation}
Proceeding as before we compute
\begin{equation}
\begin{split}
\mathbb{E}_{q(\ba)} \left[ |\bat \bx + \alpha|  \right] &= 
\frac{2}{\pi} \int_0^{\infty} \frac{dt}{t^2} \Big( 1 - e^{-\tfrac{1}{2} \xi^2 t^2} \cos ((\gamma +\alpha)t)\Big)\\
&= \Omega(\xi, \gamma + \alpha)
\end{split}
\end{equation}
and (for $\alpha_+, \alpha_->0$)
\begin{equation}
\begin{split}
&\mathbb{E}_{q(\ba)} \left[ |\bat \bx + \alpha_{+}||\bat \bx - \alpha_{-}|  \right] = \\
& \sqrt{\frac{2}{\pi}} \xi \Big\{ (\gamma+\alpha_{+}) e^{-\tfrac{1}{2} \frac{(\gamma-\alpha_{-})^2}{\xi^2}}- 
 (\gamma-\alpha_{-}) e^{-\tfrac{1}{2} \frac{(\gamma+\alpha_{+})^2}{\xi^2}} \Big\} + \\
&  \Big\{ \xi^2+(\gamma+\alpha_{+})(\gamma-\alpha_{-})\Big\}
\Big\{1-{\rm erf}\left(\frac{\gamma+\alpha_{+}}{\sqrt{2} \xi}\right) + {\rm erf}\left(\frac{\gamma-\alpha_{-}}{\sqrt{2} \xi}\right)  \Big\} \\
&\equiv \chi(\xi, \gamma, \alpha_+, \alpha_-)
\end{split}
\end{equation}
Using these identities we find:
\begin{equation}
\begin{split}
\label{eqn:phiupshardsig}
&\Phi_h= \mathbb{E}_{q(\bA)} \left[ g_\alpha(\bA \bx)_h \right] = \tfrac{1}{2}(\Omega(\xi_h, \gamma_h + \alpha)-\Omega(\xi_h, \gamma_h - \alpha)) \\
&\Upsilon_h= \mathbb{E}_{q(\bA)} \left[ g_\alpha(\bA \bx)_h^2 \right] =\tfrac{1}{2}(\xi_h^2 + \gamma_h^2 - \chi_h + \alpha^2) \qquad {\rm with} \qquad
 \chi_h \equiv \chi(\xi_h,\gamma_h,\alpha,\alpha) 
\end{split}
\end{equation}
As $\alpha\rightarrow\infty$, we have $g_\alpha(x)\rightarrow x$, i.e.~the hard sigmoid non-linearity approaches the identity function.
It is easy to verify that in this limit the expectations in Eqn.~\ref{eqn:phiupshardsig} approach the correct limit.

\subsubsection{ReLU Squared}

We consider the `ReLU squared' non-linearity, which we define to be given by 
\begin{equation}
g(x) = {\rm ReLU}(x)^2 = \Big(\tfrac{1}{2}(x + |x|)\Big)^2 = \tfrac{1}{2}(x^2 + x|x|)
\end{equation}
This is the first non-linearity we have considered that contains a piecewise quadratic portion.
Using previous results as well as the higher moments computed in the next section we find:
\begin{equation}
\begin{split}
\label{eqn:relusq}
&\Phi_h= \mathbb{E}_{q(\bA)} \left[ g(\bA \bx)_h \right] = 
\tfrac{1}{2}(\gamma_h^2 + \xi_h^2 + \Psi_h) \\
&\Upsilon_h= \mathbb{E}_{q(\bA)} \left[ g(\bA \bx)_h^2 \right] =
\tfrac{1}{2}(\gamma_h^4 + 6 \gamma_h^2 \xi_h^2  + 3 \xi_h^4 + \rho_h)
\end{split}
\end{equation}

\subsection{Higher Moments}
\label{sec:hm}

We can also compute higher moments of activation functions.
We start with the integral identities
\begin{equation}
\mathbb{E}_{q(\ba)} \left[ (\bat \bx)^2 \exp(i \bat \bx t ) \right] =\left( \xi^2 + (i \xi^2 t + \gamma)^2 \right)\exp(-\tfrac{1}{2} \xi^2 t^2 + i \gamma t) 
\end{equation}
and
\begin{equation}
\mathbb{E}_{q(\ba)} \left[ (\bat \bx)^3 \exp(i \bat \bx t ) \right] =\left\{ 3\xi^2(i \xi^2 t + \gamma) + (i \xi^2 t + \gamma)^3 \right\}\exp(-\tfrac{1}{2} \xi^2 t^2 + i \gamma t) 
\end{equation}
and
\begin{equation}
\mathbb{E}_{q(\ba)} \left[ (\bat \bx)^3  \right] = \gamma^3 + 3 \gamma \xi^2 
\end{equation}
We then have
\begin{equation}
\begin{split}
\mathbb{E}_{q(\ba)} \left[ (\bat \bx)^2 |\bat \bx|  \right] &= \frac{2}{\pi} \int_0^{\infty} \frac{dt}{t^2} 
\Big( \xi^2 + \gamma^2 - (\xi^2+\gamma^2-\xi^4t^2) e^{-\tfrac{1}{2} \xi^2 t^2} \cos (\gamma t) + 2\xi^2 \gamma t e^{-\tfrac{1}{2} \xi^2 t^2} \sin (\gamma t)\Big)\\
&= (\xi^2 + \gamma^2) \left\{  \frac{2}{\pi} \int_0^{\infty} \frac{dt}{t^2} \Big( 1 - e^{-\tfrac{1}{2} \xi^2 t^2} \cos (\gamma t)\Big) \right\}  
+\frac{2\xi^4}{\pi} \int_0^{\infty} \!\! dt e^{-\tfrac{1}{2} \xi^2 t^2} \cos (\gamma t)  \\
&+2\xi^2 \gamma \frac{\partial}{\partial \gamma} \left\{\frac{2}{\pi} \int_0^{\infty} \frac{dt}{t^2} \Big( 1 - e^{-\tfrac{1}{2} \xi^2 t^2} \cos (\gamma t)\Big) \right\}\\
&= (\xi^2 + \gamma^2) \left\{ \sqrt{\frac{2}{\pi}} \xi e^{-\tfrac{1}{2} \frac{\gamma^2}{\xi^2}} + \gamma {\rm erf}\left(\frac{\gamma}{\sqrt{2} \xi}\right)\right\} 
+ \sqrt{\frac{2}{\pi}}\xi^3 e^{-\tfrac{1}{2} \frac{\gamma^2}{\xi^2}} \\
&+ 2\xi^2 \gamma \frac{\partial}{\partial \gamma} \left\{ \sqrt{\frac{2}{\pi}} \xi e^{-\tfrac{1}{2} \frac{\gamma^2}{\xi^2}} + \gamma {\rm erf}\left(\frac{\gamma}{\sqrt{2} \xi}\right)\right\}\\
&= \sqrt{\frac{2}{\pi}}(2\xi^3 + \gamma^2 \xi) e^{-\tfrac{1}{2} \frac{\gamma^2}{\xi^2}} + (3\gamma\xi^2 + \gamma^3) {\rm erf}\left(\frac{\gamma}{\sqrt{2} \xi}\right) \\
&\equiv \zeta(\xi, \gamma)
\end{split}
\end{equation}
as well as 
\begin{equation}
\begin{split}
\mathbb{E}_{q(\ba)} \left[ (\bat \bx)^3 |\bat \bx|  \right] &= \frac{2}{\pi} \int_0^{\infty} \frac{dt}{t^2} 
\Big( \gamma^3 + 3\gamma \xi^2 - (\gamma^3+3\gamma\xi^2-3\gamma\xi^4t^2) e^{-\tfrac{1}{2} \xi^2 t^2} \cos (\gamma t) \\
&+ (3\xi^4 t + 3\gamma^2 \xi^2t + \xi^6 t^3) e^{-\tfrac{1}{2} \xi^2 t^2} \sin (\gamma t)\Big)\\
&= ( \gamma^3 + 3\gamma \xi^2) \left\{  \frac{2}{\pi} \int_0^{\infty} \frac{dt}{t^2} \Big( 1 - e^{-\tfrac{1}{2} \xi^2 t^2} \cos (\gamma t)\Big) \right\}  
+\frac{6\gamma\xi^4}{\pi} \int_0^{\infty} \!\! dt e^{-\tfrac{1}{2} \xi^2 t^2} \cos (\gamma t)  \\
&+(3\xi^4  + 3\gamma^2 \xi^2)  \frac{\partial}{\partial \gamma} \left\{\frac{2}{\pi} \int_0^{\infty} \frac{dt}{t^2} \Big( 1 - e^{-\tfrac{1}{2} \xi^2 t^2} \cos (\gamma t)\Big) \right\}\\
&-2\xi^6  \frac{\partial}{\partial \xi^2}  \frac{\partial}{\partial \gamma} \left\{\frac{2}{\pi} \int_0^{\infty} \frac{dt}{t^2} \Big( 1 - e^{-\tfrac{1}{2} \xi^2 t^2} \cos (\gamma t)\Big) \right\}\\
&= \sqrt{\frac{2}{\pi}} (\gamma^3 \xi + 7\gamma \xi^3) e^{-\tfrac{1}{2} \frac{\gamma^2}{\xi^2}} + (\gamma^4+6\gamma^2\xi^2 + 3\xi^4) {\rm erf}\left(\frac{\gamma}{\sqrt{2} \xi}\right) \\
&\equiv \rho(\xi, \gamma)
\end{split}
\end{equation}

\subsection{Piecewise Polynomial Activation Functions}

Piecewise polynomial functions in $x$ can be represented by composing polynomials in $x$ with the absolute value function.
Thus in order to compute Eqn.~\ref{eqn:phiupsdef} for general piecewise polynomial activation functions we need to be able to compute expectations of the form 
\begin{equation}
\mathbb{E}_{q(\ba)} \left[ p_0(\bat \bx) \prod_i |p_i(\bat \bx)|  \right] 
\end{equation}
where the $p_i(\cdot)$ are polynomials. We have shown how this computation can be done in a number of cases. For any specific case the recipe we have used to do the computation remains applicable. In particular one can compute any needed `base' integrals by doing the computation in one dimension as in Eqn.~\ref{eqn:relumean2}. One can then make use of the integral identity in Eqn.~\ref{eqn:absrep} and differentiation to compute higher order moments via purely algebraic operations (c.f.~the manipulations in Sec.~\ref{sec:hm}).

\subsection{Other likelihoods}
\label{sec:otherlike}

In the main text we showed how to compute exact closed form expressions for the ELBO variational objective
in the regression case. For other likelihoods, the required expectations are generally intractable. Nevertheless we can still compute
closed form variational objectives at the price of some approximation. Alternatively, if we are worried about the bias introduced by our approximations, we can use our approximations as control variates in the Monte Carlo sampling setting. In this section we briefly describe how this goes in the case of softmax classification.

\subsubsection{Softmax Categorical Likelihood}
\label{sec:softmax}

Using familiar bounds\footnote{See e.g.~\texttt{http://www.columbia.edu/{\textasciitilde}jwp2128/Teaching/E6720/Fall2016/papers/twobounds.pdf}} we have:
\begin{equation}
\begin{split}
\mathbb{E}_{q(\bA)q(\bB)} \left[\log p(y=k|\bA, \bB, \bx) \right] &= 
\mathbb{E}_{q(\bA)q(\bB)} \left[\log \frac{e^{\by_k}}{\sum_j e^{\by_j}} \right] \\
&\ge \mathbb{E}_{q(\bA)q(\bB)} \left[ \by_k \right] - \log \sum_j \mathbb{E}_{q(\bA)q(\bB)} \left[ e^{\by_j} \right]
\end{split}
\end{equation}
We do a second-order Taylor expansion of $\by_j$ around its expectation $\byt_j$ to obtain
\begin{equation}
\begin{split}
\mathbb{E}_{q(\bA)q(\bB)} \left[ e^{\by_j} \right] &\approx e^{\byt_j} \mathbb{E}_{q(\bA)q(\bB)} \left[1+ (\by_j - \byt_j) + \frac{1}{2}(\by_j - \byt_j)^2 \right] \\
&=e^{\byt_j}\Big(1 + \tfrac{1}{2}  {\rm var}(\by_j)\Big)
\end{split}
\end{equation}
so that our approximate lower bound to the expected log likelihood becomes
\begin{equation}
\begin{split}
\mathbb{E}_{q(\bA)q(\bB)} \left[\log p(y=k|\bA, \bB, \bx) \right] \gtrapprox
\byt_k -  \log \sum_j e^{\byt_j}\Big(1 + \tfrac{1}{2}  {\rm var}(\by_j)\Big)
\end{split}
\label{sec:softmaxapprox}
\end{equation}
We can then use the closed form expressions for the mean function and variance given in the main text to form a deterministic approximation to the expected log likelihood.

\subsubsection{Logistic Bernoulli Likelihood}
For the case with two classes with $y\in \{0,1\}$ and where we have a single logit $\hat{y}$ the approximation in
Eqn.~\ref{sec:softmaxapprox} reduces to
\begin{equation}
\begin{split}
\mathbb{E}_{q(\bA)q(\bB)} \left[\log p(y|\bA, \bB, \bx) \right] \gtrapprox
y \hat{y} -  \log \Big( 1 + e^{\hat{y}}\left[1 + \tfrac{1}{2}  {\rm var}(\hat{y})\right]\Big)
\end{split}
\end{equation}

\subsubsection{Control Variates}

To reduce bias to zero one can construct a control variate \citep{ross} version of the variational objective in Eqn.~\ref{sec:softmaxapprox}:
\begin{equation}
\begin{split}
&\mathbb{E}_{q(\bA)q(\bB)} \left[\log p(y=k|\bA, \bB, \bx) \right] = \\
& \byt_k -  \log \sum_j e^{\byt_j}\Big(1 + \tfrac{1}{2}  {\rm var}(\by_j)\Big)-
\mathbb{E}_{q(\bA)q(\bB)} \left[ \log \sum_j e^{\by_j} \right]  +\\
& \log \Big( \sum_j e^{\byt_j} \mathbb{E}_{q(\bA)q(\bB)} \left[ 1+ (\by_j - \byt_j) + \frac{1}{2}(\by_j - \byt_j)^2 \right] \Big)
\end{split}
\label{sec:softmaxcv}
\end{equation}
The expectations in Eqn.~\ref{sec:softmaxcv} are then estimated with Monte Carlo, while the rest is
available in closed form. Alternatively, we can use the following estimator:
\begin{equation}
\begin{split}
&\mathbb{E}_{q(\bA)q(\bB)} \left[\log p(y=k|\bA, \bB, \bx) \right] = 
\byt_k - \mathbb{E}_{q(\bA)q(\bB)} \left[ \log \sum_j e^{\by_j} \right] 
\end{split}
\label{sec:approx2}
\end{equation}
That is, we use the analytic result for the numerator in the softmax likelihood and sample the troublesome denominator.

\subsubsection{Fast Approximate Prediction}
\label{sec:fastpred}

To make fast test time predictions 
we can simply use the analytic mean function $\byt$, i.e.~use 
\begin{equation}
{\rm pred}(\bx) = {\rm argmax}_k \byt_k(\bx)
\end{equation}
effectively ignoring the 
normalizing term in the softmax likelihood. As can be seen in Fig.~\ref{fig:resnet}, this approximation can be quite
effective in practice.

\subsection{Experimental Details}

All the experiments described in this work were implemented in the Pyro probabilistic programming language \citep{bingham2018pyro},
which is built on top of PyTorch \citep{paszke2017automatic}. As noted in the main text, whenever sampling a weight matrix we make use of the `local reparameterization trick' \citep{kingma2015variational}, i.e.~we sample in pre-activation space and not in weight space. This can lead to substantial variance reduction as compared to sampling in weight space directly.

\subsubsection{Variance Reduction}

We use the 90-dimensional `YearPredictionMSD' dataset from the UCI repository \citep{Dua:2017}.
This dataset is a subset of the Million Song Dataset \citep{Mahieux2011}. The architecture of our neural
network is given by $90 - 200 - 200 - 1$,
where all layers are fully connected and both non-linearities are ReLU. We use mean field (Normal) variational distributions
for all weight matrices. To compute gradient variances of variational
parameters with respect to the variational objective,
we fix a random mini-batch of training data with 500 elements. We then compute $10^4$ samples and report
empirical gradient variances averaged over the elements of each tensor. We report gradient variances computed before any training as well as after 50 epochs. For the (partially) analytic result, only the weight matrix closest to the inputs is sampled, while for the sampling result all weight matrices are sampled.

\subsubsection{VAE with a Bayesian Decoder}

We use the same dataset as for the variance reduction experiment above (with the difference that in this unsupervised setting we only use the input features). 
This dataset has $N=515345$ data points.
We split the data into training, test, and validation sets in the proportion 7 : 2 : 1. For the encoder we use a fully connected
(non-Bayesian) neural network with 500 hidden units in each of the two hidden layers.
 For the decoder we use a neural network with
a single hidden layer with 500 hidden units. All non-linearities are ReLU. We use the Adam optimizer \citep{kingma2014adam}
and mini-batches of size 2000 during training. We use mean field (Normal) variational distributions
for all weight matrices in the decoder. We do a grid search over the hyperparameters of the optimizer and use the validation set to choose the number of epochs to train. For all three model variants this procedure resulted in the following choices: default Adam hyperparameters and 1500 epochs of training. We use a latent dimension of 30. We report test log likelihoods that make use of an importance weighted estimator that draws 500 $\times$ 100 samples per data point (100 samples inside the log averaged over 500 trials).

\subsubsection{Fast Prediction}

We take a ResNet50 \citep{he2016deep} that is pre-trained\footnote{\texttt{https://pytorch.org/docs/stable/torchvision/models.html}} on ImageNet \citep{russakovsky2015imagenet}
and then lop off the final layer and replace it with the following neural network architecture: $2048 - 1000 - 1000 - 1000$.
Here the first two dashes represent ReLU non-linearities and the final layer of outputs represents softmax logits. We learn
the first weight matrix (2048 - 1000) using MLE and are Bayesian about the subsequent weight matrices (we use mean 
field variational distributions).
We do not fine-tune the weight matrices inherited from the pre-trained ResNet50. Our test set and validation set consist of
40k and 10k images, respectively. We train\footnote{Note that we also trained our network using the approximations and control variates described in Sec.~\ref{sec:otherlike}, but we do not report those results here.} for up to 120 epochs and use the validation set to fix optimization hyperparameters and determine how many epochs to train. 
In contrast to the typical approach taken in deep learning, we used a fixed size/crop for training images, i.e.~we do not do any data augmentation. 

Fig.~\ref{fig:resnet} is generated as follows. For the MC estimate of the predicted class probabilities, we draw a total of 4096 samples per datapoint. These samples are then combined via the following allocation:
\begin{equation}
N_{\rm inner} \in [1, 2, 4, 8, 16, 32, 64, 128, 256, 512] \qquad {\rm and} \qquad N_{\rm outer} = 4096 / N_{\rm inner}
\end{equation}
Here $N_{\rm inner}$, which represents the number of samples inside the log, is the quantity plotted on the horizontal
axis of Fig.~\ref{fig:resnet}.
To form the deterministic approximation we follow Sec.~\ref{sec:fastpred}.

\subsection{Related Work}

The approach most closely related to ours is probably the deterministic approximations in ref.~\citep{wu2018fixing} (indeed they compute some of the same ReLU integrals that we do). While we focus on single-layer neural networks, the distinct advantage of their approximation scheme is that it can be applied to networks of arbitrary depth. Thus some of our results are potentially complementary to theirs. Reference \citep{kandemir2018sampling} also constructs deterministic variational objectives for the specific case of the ReLU activation function. 
Reference \cite{marlin2011piecewise} considers quadratic piecewise linear bounds for the logistic-log-partion function in the context of Bernoulli-logistic latent Gaussian models.
Finally, approaches for variance reduction in the stochastic variational inference setting include \citep{kingma2015variational} and \citep{wen2018flipout}.

\subsection{Assorted Remarks}

\begin{enumerate}
\item The factorization assumption in Eqn.~\ref{eqn:varform} can probably be weakened at the cost of dealing with special functions more exotic than the error function.
\item Note that the expression for the full (predictive) variance, which decomposes into three readily identified components, is
given by:
\begin{equation}
{\rm var}(\bx)= 
 \underbrace{{\rm diag}(\bXi_\bB) \cdot (\bUps - \bPhi \odot \bPhi)}_{\rm variance \;from\; \bA} \;\;+  
 \underbrace{\bPhi^{\rm T}\bSig_\bB\bPhi}_{\rm variance \;from\; \bB} +
 \underbrace{\beta^{-1}}_{\rm observation\; noise}
\end{equation}
\item Although we do not do so here, it would probably be straightforward to compute an all orders formula for the moments of the ReLU activation function, for which the main computational ingredient is the expectation
\begin{equation}
\mathbb{E}_{q(\ba)} \left[ (\bat \bx)^n |\bat \bx|  \right]
\end{equation}
\end{enumerate}

\end{document}